\documentclass{article}
\usepackage{spconf,amsmath,graphicx}

\usepackage{enumitem}
\setlist{nosep, leftmargin=14pt}

\usepackage{mwe} 
\usepackage[hyphens]{url}
\usepackage{siunitx}
\usepackage{amsmath}
\usepackage{amssymb}
\usepackage{dsfont}
\usepackage{booktabs}
\usepackage{multirow}  
\usepackage{stmaryrd} 
\usepackage{hyperref}

\newcommand{\eg}{e.g., }
\newcommand{\RNum}[1]{\uppercase\expandafter{\romannumeral #1\relax}}

\title{Stylizing ViT: Anatomy-Preserving Instance Style Transfer for Domain Generalization}
\name{Sebastian Doerrich \qquad Francesco Di Salvo \qquad Jonas Alle \qquad Christian Ledig}
\address{xAILab Bamberg, University of Bamberg, Bamberg, Germany}
\begin{document}
\ninept
\maketitle
\begin{abstract}
Deep learning models in medical image analysis often struggle with generalizability across domains and demographic groups due to data heterogeneity and scarcity. Traditional augmentation improves robustness, but fails under substantial domain shifts. Recent advances in stylistic augmentation enhance domain generalization by varying image styles but fall short in terms of style diversity or by introducing artifacts into the generated images. To address these limitations, we propose \textit{Stylizing ViT}, a novel Vision Transformer encoder that utilizes weight-shared attention blocks for both self- and cross-attention. This design allows the same attention block to maintain anatomical consistency through self-attention while performing style transfer via cross-attention.
We assess the effectiveness of our method for domain generalization by employing it for data augmentation on three distinct image classification tasks in the context of histopathology and dermatology.
Results demonstrate an improved robustness (up to +\qty{13}{\%} accuracy) over the state of the art while generating perceptually convincing images without artifacts. Additionally, we show that \textit{Stylizing ViT} is effective beyond training, achieving a \qty{17}{\%} performance improvement during inference when used for test-time augmentation. The source code is available at \href{https://github.com/sdoerrich97/stylizing-vit}{https://github.com/sdoerrich97/stylizing-vit} .
\end{abstract}
\begin{keywords}
style transfer, domain generalization, cross-attention, data augmentation, vision transformer
\end{keywords}
\section{Introduction}
\label{sec:introduction}
\begin{figure}[htb]
    \centering
    \includegraphics[width=0.93\linewidth]{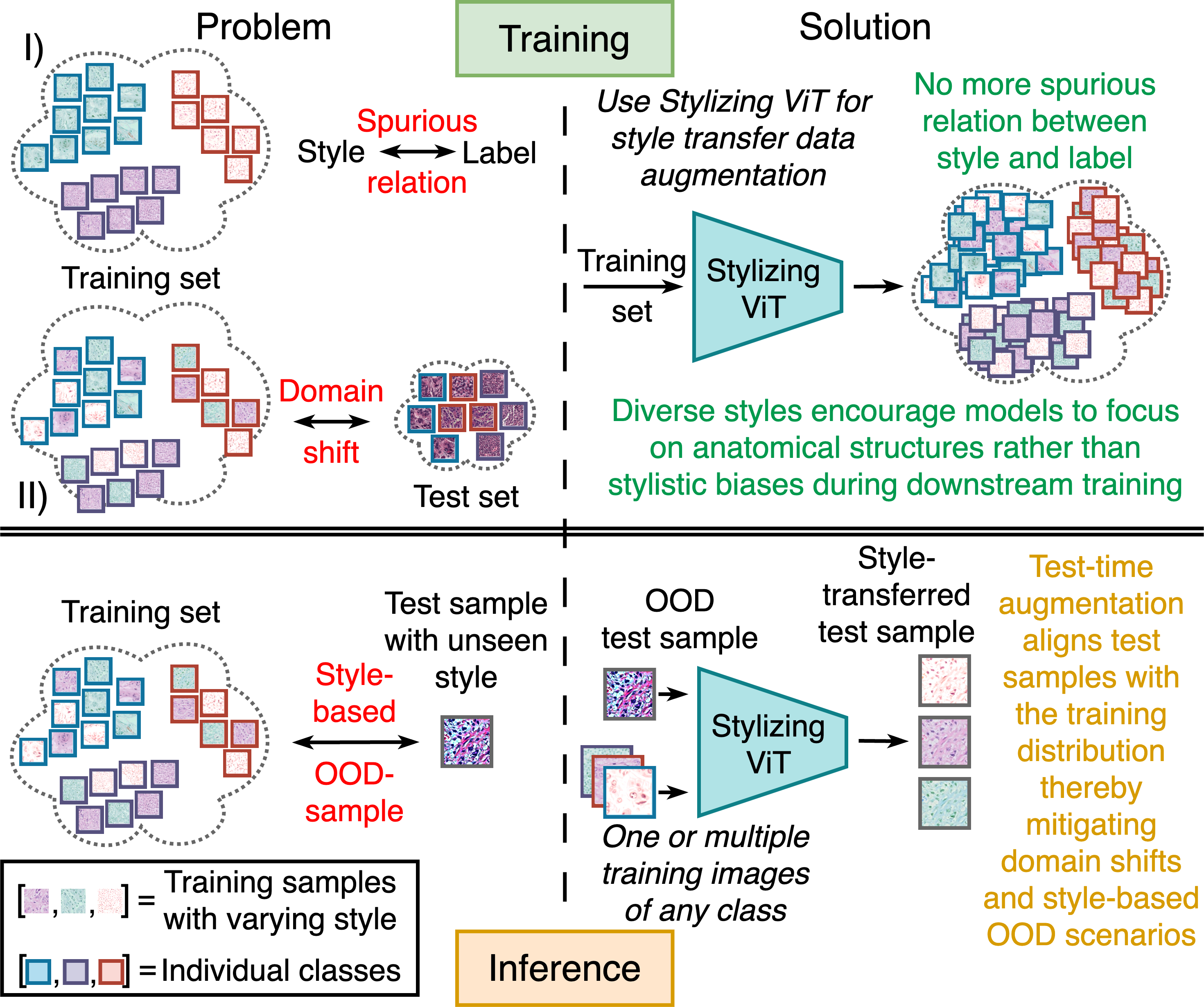}
    \caption{Overview of how \textit{Stylizing ViT} is used to improve domain generalization during training and inference. During training, it generates stylistically diverse but anatomically consistent images, encouraging downstream classifiers to learn structure-aware representations. At test time, it is used to align unseen input styles with the training distribution, thereby mitigating style-induced domain shifts.}
    \label{fig:pullFigure}
\end{figure}
Deep learning has achieved remarkable success in image analysis, text processing, and even computer assisted interventions, yet its integration into clinical practice remains limited \cite{Stacke2021}.
A major challenge in medical applications stems from the diverse, heterogeneous, and inherently scarce nature of medical data~\cite{Khan2022,Schafer2024}, which affects the generalizability of deep learning models across institutions, domains, and demographic groups~\cite{Dulaney2024}.
Traditional data augmentation techniques, particularly those utilizing modality-specific characteristics like color gradients or artifacts, improve model robustness by diversifying training data but struggle with substantial domain shifts~\cite{Islam2024,Ouyang2023,Yoon2024}. In response, recent work suggests that stylistic data augmentation enhances domain generalization by altering image styles~\cite{doerrich2024,nguyen2024}.
However, these methods either offer limited style diversity or introduce artifacts into the generated images.

In this work, we address these challenges by building on advances in neural style transfer. Neural style transfer aims to render a natural image in the appearance of an artwork while preserving its underlying structure. The seminal work of Gatys et al.~\cite{Gatys2015ANA} introduced this concept using Gram-matrix–based optimization, but its iterative nature made it computationally expensive and slow. This limitation motivated the development of real-time, arbitrary style transfer methods such as AdaIN~\cite{huang2017} and AdaAttN~\cite{liu2021} where any style can be transferred without having to retrain the model. While early approaches relied on convolutional neural networks (CNNs), recent works increasingly adopt Transformers for their reduced content bias and superior capacity to model long-range dependencies~\cite{deng2022}. However, most methods still require a combination of two encoders and a decoder, resulting in a considerable computational overhead.
\begin{figure*}[htb]
    \centering
    \includegraphics[width=0.99\linewidth]{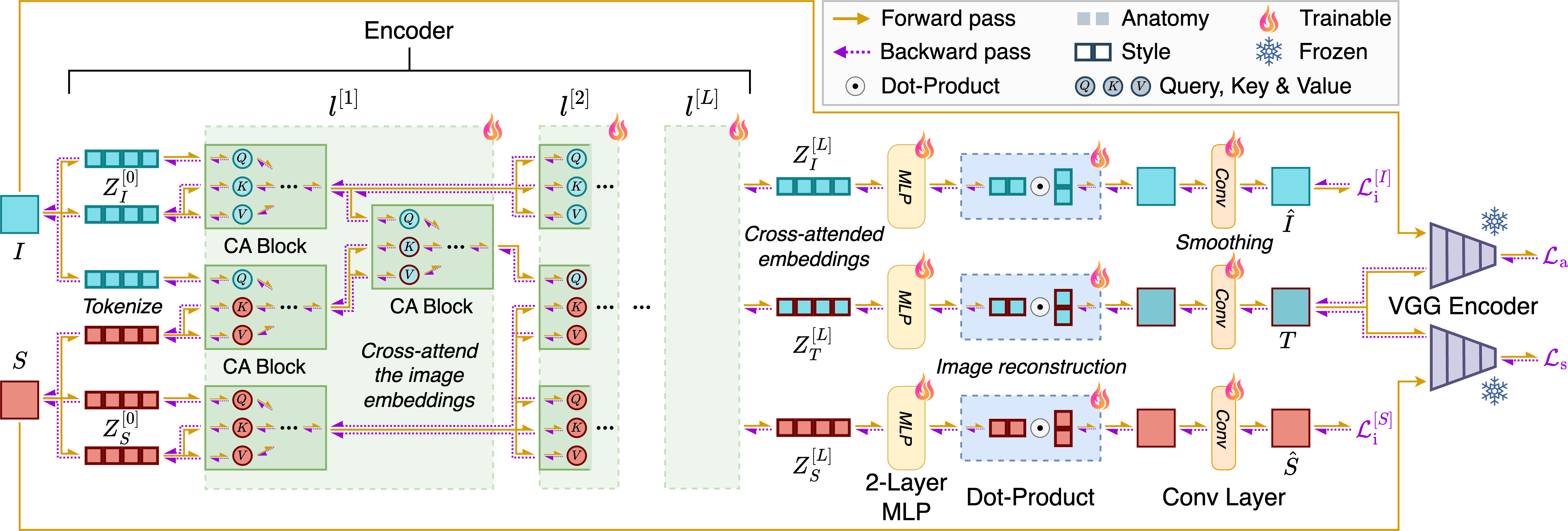}
    \caption{Overview of our proposed \textit{Stylizing ViT}. The input image pair ($I$, $S$) is processed by the encoder that fuses the anatomical structure of $I$ with the style characteristics of $S$ using cross-attention. Subsequently the stylized image $T$ is reconstructed through a two-layer MLP, a dot-product operation, and a convolutional layer. A frozen VGG19 encoder is used during training to compute perceptual losses.}
    \label{fig:method}
\end{figure*}

To tackle these limitations, we introduce \textit{Stylizing ViT}, a novel style transfer method based on a single-encoder Vision Transformer (ViT). Our architecture employs weight sharing within a unified attention block to jointly fuse self- and cross-attention, enabling effective style transfer while preserving anatomical fidelity.
We show that \textit{Stylizing ViT} generates realistic, anatomically accurate images with diverse styles across dataset splits and domains. This makes it an effective augmentation technique for augmenting training images on-the-fly, thereby encouraging models to focus on anatomical structures rather than stylistic biases during downstream training. We evaluate this aspect of \textit{Stylizing ViT} across two histopathology datasets and a dermatology one, showing improved performance across domains including varying staining techniques and skin tones. Furthermore, we demonstrate that our approach is able to generalize to unseen anatomical and stylistic variations, highlighting its applicability beyond training-time augmentation, \eg for inference. \figurename~\ref{fig:pullFigure} illustrates these two key application scenarios side-by-side. Our main contributions include:
\begin{itemize}
    \item A modality-agnostic style augmentation method for domain generalization.
    \item A novel ViT design enabling weight sharing within a unified attention block for both self- and cross-attention information fusion, termed \textit{Stylizing ViT}.
    \item Applicability for test-time augmentation (TTA) to enhance generalization to new, unseen domains during inference.
    \item Comprehensive experiments across three medical imaging datasets, demonstrating consistent improvements in style transfer quality, state-of-the-art classification performance (up to +\qty{9}{\%} accuracy over prior best), and significant gains from 
    TTA.
\end{itemize}
\section{Method}
The overview of our method is illustrated in \figurename~\ref{fig:method}. Given two input images, $I$ and $S$, the objective is to transfer the style of $S$ onto $I$ while preserving $I$'s anatomical integrity. This is achieved through a sequence of model components, including a ViT encoder, a two-layer MLP, a dot-product operation on the resulting feature embeddings, and a final convolutional layer.
Specifically, $I$ and $S$ are first processed by a single, shared ViT encoder, where all self-attention operations are replaced with cross-attention to enable the alignment of $I$'s anatomical structure with the style of $S$. The subsequent MLP scales the feature embeddings for the image reconstruction via the dot-product which restores the original image dimensions. Finally, a convolutional layer is applied to learn more consistent outputs. By combining cross-attention for stylization with robust loss functions from a frozen VGG19 encoder, our method achieves high-quality stylization results without requiring a decoder for reconstruction.
\subsection{Stylizing ViT for Style Transfer}
We utilize the structural similarity between cross-attention (CA) and self-attention (SA) to introduce \textit{Stylizing ViT}, a Vision Transformer~\cite{Dosovitskiy2021} variant that enables the stylization of an image $I$ with the style of a reference image $S$ to produce the style transferred image $T$. Specifically, we modify the standard ViT encoder by replacing all SA operations with CA, allowing for controlled fusion of anatomical and stylistic information. Furthermore, this allows to employ the same, weight-shared attention block either in the mode of SA or CA. Thus, a single block first processes each input image separately via its SA mode to preserve distinct structural and stylistic representations for the next layer while simultaneously the same block is used to blend the anatomical structure of $I$ with the style attributes of $S$ through CA.
Specifically, for tokenized embeddings $Z_{I}$ and $Z_{S}$ of $I$ and $S$, each encoder layer $l$ applies the same, weight-shared CA block in three stages: (i) CA($Z_{I}, Z_{I}$) which is equivalent to SA($Z_{I}$), (ii) CA($Z_{S}, Z_{S}$) which is equivalent to SA($Z_{S}$), and (iii) CA($Z_{I}, Z_{S}$) to produce the stylized output $Z_{T}$. A final CA step, using still the same, weight-shared block, further aligns $Z_{I}$ and $Z_{T}$ for $l < L$ to ensure anatomical preservation during the stylization.
\subsection{Image Reconstruction}
To reconstruct the original images $I$ and $S$ as well as the stylized image $T$, the output embeddings $Z_{I}$, $Z_{S}$, and $Z_{T}$ from the encoder are first processed by a shared two-layer MLP. Each embedding is then split into two halves ($z_\text{A}, z_\text{B} \in \mathds{R}^{n \times d/2}$) along embedding dimension $d$, shaped into matrix form ($Z_\text{A} \in \mathds{R}^{n \times c \times p \times v}$, $Z_\text{B} \in \mathds{R}^{n \times c \times v \times p}$), and multiplied via dot product multiplication along $v$.
Here, $n$ represents the number of image patches, $c$ the channel dimension, $p$ the patch size, and $v = d / (c \times p)$ the hidden dimension for the dot product. To restore the original image dimension the initial patchification from the ViT encoder is reversed. The final convolutional layer with a kernel size of five is used to learn more consistent reconstructions.
\subsection{Network Optimization}
\label{subsec:optimization}
\begin{table*}[htb]
\caption{Quantitative evaluation of reconstruction and style transfer quality on the full training sets of Camelyon17-WILDS, Epithelium-Stroma, and Fitzpatrick17k. Reconstruction quality is assessed using PSNR and SSIM on identical image pairs while style transfer quality is evaluated using FID, LPIPS, and ArtFID on distinct image pairs. Higher values ($\shortuparrow$) indicate better performance for PSNR and SSIM, while lower values ($\shortdownarrow$) are better for FID, LPIPS, and ArtFID. The \textbf{best performance} for each metric is highlighted in bold.}
\label{tab:trainingSetStylization}
    \centering
    \small
    \setlength{\tabcolsep}{1.1mm}
    \begin{tabular}{l r r r r r r r r r r r r r r r r r}
        \toprule
        \multirow{3.5}{*}{Method} & \multicolumn{5}{c}{Camelyon17-WILDS} & & \multicolumn{5}{c}{Epithelium-Stroma} & & \multicolumn{5}{c}{Fitzpatrick17k} \\
        & \multicolumn{2}{c}{Reconstruct $\shortuparrow$} & \multicolumn{3}{c}{Style Transfer $\shortdownarrow$} & & \multicolumn{2}{c}{Reconstruct $\shortuparrow$} & \multicolumn{3}{c}{Style Transfer $\shortdownarrow$} & & \multicolumn{2}{c}{Reconstruct $\shortuparrow$} & \multicolumn{3}{c}{Style Transfer $\shortdownarrow$} \\
        \cmidrule(lr){2-3} \cmidrule(lr){4-6} \cmidrule(lr){8-9} \cmidrule(lr){10-12} \cmidrule(lr){14-15} \cmidrule(lr){16-18}
        & PSNR & SSIM & FID & LPIPS & ArtFID & & PSNR & SSIM & FID & LPIPS & ArtFID & & PSNR & SSIM & FID & LPIPS & ArtFID \\
        \midrule
        AdaIN         & 23.7 & 0.75 & 55.9  & 0.17 & 66.6  & & 23.6 & 0.85 & 22.9  & 0.13 & 26.9   & & 20.4 & 0.61 & 43.5  & 0.21 & 54.0  \\
        IEContrAST    & 27.4 & 0.82 & 138.7 & 0.33 & 185.4 & & 29.3 & 0.91 & 36.4  & 0.22 & 45.4   & & 30.4 & 0.77 & 60.9  & 0.39 & 85.9  \\
        SANet         & 26.8 & 0.81 & 113.8 & 0.29 & 148.6 & & 30.4 & 0.91 & 35.1  & 0.21 & 43.8   & & 29.4 & 0.76 & 81.4  & 0.44 & 118.7 \\
        StyTR2        & 30.0 & 0.88 & 84.1  & 0.29 & 109.4 & & 30.8 & 0.92 & 28.4  & 0.20 & 35.2   & & 29.4 & 0.75 & 104.5 & 0.47 & 154.8 \\
        SGViT & \textbf{46.7} & \textbf{0.99} & 172.4 & 0.34 & 232.7 & & \textbf{47.6} & \textbf{0.99} & 175.6 & 0.49 & 262.92 & & \textbf{40.5} & \textbf{0.97} & 226.0 & 0.54 & 348.8 \\
        \midrule    
        Stylizing ViT & 45.4 & \textbf{0.99} & \textbf{6.2}   & \textbf{0.06} & \textbf{7.6}   & & 39.0 & \textbf{0.99} & \textbf{1.5}   & \textbf{0.02} & \textbf{2.6}    & & 31.5 & 0.77 & \textbf{36.9}  & \textbf{0.20} & \textbf{45.5}  \\
        \bottomrule
    \end{tabular}
\end{table*}
The stylized image $T$ should preserve the anatomical structures of $I$ while incorporating the style patterns of $S$. To achieve this, we employ a strategy common in neural style transfer methods~\cite{huang2017,liu2021,deng2022}. Specifically, we define two perceptual loss terms: an anatomy loss $\mathcal{L}_{\text{a}}$ that quantifies structural differences between $T$ and $I$, and a style loss $\mathcal{L}_{\text{s}}$ that measures stylistic discrepancies between $T$ and $S$. Additionally, we incorporate the consistency loss $\mathcal{L}_{\text{c}}$ and identity loss $\mathcal{L}_{\text{i}}$ from~\cite{deng2022} to guide the network in learning accurate representations. All loss terms, except for $\mathcal{L}_{\text{i}}$, are computed using intermediate feature maps $k \in K$ extracted from a frozen, ImageNet-pretrained VGG19 model, aligning with~\cite{Gatys2015ANA}. The individual loss functions $\mathcal{L}_{\text{i}}, \mathcal{L}_{\text{c}}, \mathcal{L}_{\text{a}}, \mathcal{L}_{\text{s}}$ as well as the total loss $\mathcal{L}_{\text{t}}$ can be written as:
\begin{equation*}
    \begin{aligned}
        \mathcal{L}_{\text{i}} &= \|I - \hat{I}\|_2 + \|S - \hat{S}\|_2
        \\
        \mathcal{L}_{\text{c}} &= \sum \| \phi_k(I) - \phi_k(\hat{I}) \|_2 + \| \phi_k(S) - \phi_k(\hat{S})\|_2
        \\
        \mathcal{L}_{\text{a}} &= \sum \| \phi_k(I) - \phi_k(T) \|_2 
        \\
        \mathcal{L}_{\text{s}} &= \sum \| \mu(\phi_k(S)) - \mu(\phi_k(T)) \|_2  + \| \sigma(\phi_k(S)) - \sigma(\phi_k(T)) \|_2
        \\
        \mathcal{L}_{\text{t}} &= \lambda_{\text{i}} \mathcal{L}_{\text{i}} + \lambda_{\text{c}} \mathcal{L}_{\text{c}} + \lambda_{\text{a}} \mathcal{L}_{\text{a}} + \lambda_{\text{s}} \mathcal{L}_{\text{s}}
    \end{aligned}
\end{equation*}
where $\hat{I}$ and $\hat{S}$ represent the reconstructions of $I$ and $S$, respectively, $\phi_k(\cdot)$ the feature embeddings from the $k$-th layer, and $\mu(\cdot)$ and $\sigma(\cdot)$ the mean and standard deviation of the extracted features. The weights $\lambda_{\text{i}}, \lambda_{\text{c}}, \lambda_{\text{a}}, \lambda_{\text{s}}$ are set to $70$, $1$, $7$, $10$, in alignment with~\cite{deng2022}.

For our \textit{Stylizing ViT}, we adopt a ViT-B/16 backbone and train it independently on each dataset’s training split for $50$ epochs with a batch size of $64$ using AdamW (learning rate of $0.001$) and a cosine annealing schedule. To obtain distinct anatomy-style pairs, we employ a self-supervised approach: the same image batch is loaded twice, with the first batch being used for the anatomy images while the samples in the second batch, the style images, are shuffled to avoid corresponding anatomy-style pairs of identical samples.
\section{Experiments and Results}
\label{sec:experimentsAndResults}
We evaluate our method for (1) its effectiveness in transferring styles while preserving anatomical structures, (2) its utility as a data augmentation technique for enhancing domain generalization, and (3) its potential for TTA.
Our evaluation employs two histopathology and one dermatology dataset. The first, Camelyon17-WILDS~\cite{wilds2021}, comprises H\&E-stained lymph node image patches for tumor identification across five hospitals (H1-H5). We adopt the official dataset splits: train (H1-H3; 302,436 samples), val (H4; 34,904), and test (H5; 85,054).
The second histopathology dataset, sourced from~\cite{Li2022}, aggregates three public datasets for epithelium-stroma classification, namely NKI~\cite{Beck2011} (train: 8,337), VGH~\cite{Beck2011} (val: 5,920), and IHC~\cite{Linder2012} (test: 1,376). NKI and VGH contain H\&E-stained breast cancer images, while IHC consists of IHC-stained colorectal cancer images.
Finally, we utilize Fitzpatrick17k~\cite{Groh2021}, a dataset of dermatology images labeled with Fitzpatrick skin tones (\RNum{1}-\RNum{6}). Following~\cite{Daneshjou2021D}, we define three groups: \{\RNum{1}-\RNum{2}: train (7,755), \RNum{3}-\RNum{4}: val (6,089), \RNum{5}-\RNum{6}: test (2,168)\}, representing progressively darker skin tones. All datasets are standardized to a $224 \times 224$ resolution using bicubic interpolation while maintaining aspect ratios.
\subsection{Training Set Stylization}
\label{subsec:TrainingSetStylizationResults}
To evaluate the effectiveness of our method in transferring style while preserving anatomical fidelity, we assess both reconstruction and style transfer quality on the full training sets. For reconstruction, we compute the Peak Signal-to-Noise Ratio (PSNR) and Structural Similarity Index Measure (SSIM) which quantify the preservation of fine-grained structural detail on pairs of identical images. For style transfer evaluation, we use pairs of distinct images and compute the Fr\'echet Inception Distance (FID)~\cite{Heusel2017}, Learned Perceptual Image Patch Similarity (LPIPS)~\cite{Zhang2018TheUE}, and ArtFID~\cite{Wright2022ArtFID}. FID measures how closely the stylized image matches the style distribution of the style image, LPIPS quantifies content fidelity between the stylized and content image, and ArtFID jointly captures both content preservation and stylistic accuracy, aligning well with human judgment~\cite{Chung2024}.

We compare our method against four state-of-the-art neural style transfer methods: AdaIN~\cite{huang2017}, IEContrAST~\cite{Chen2021IEContrAST}, SANet~\cite{Park2018ArbitraryST}, and $\text{StyTR}^\text{2}$~\cite{deng2022} as well as the recent SGViT~\cite{doerrich2024}. All of these were trained per dataset using their official implementations and recommended training settings.

\begin{figure}[htb]
    \centering
    \includegraphics[width=\linewidth]{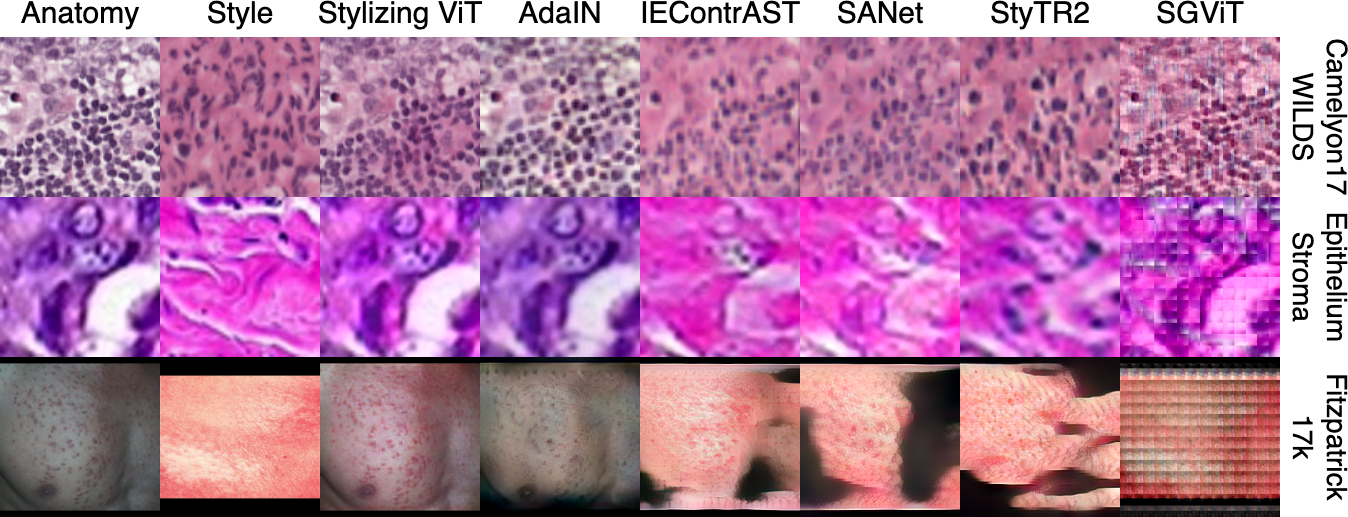}
    \caption{Qualitative comparison of style transfer quality on training image pairs. Our method consistently outperforms all reference methods in applying diverse styles without compromising anatomy.}
    \label{fig:stylizedImages}
\end{figure}

Quantitative results are presented in Table~\ref{tab:trainingSetStylization}, with representative qualitative examples shown in \figurename~\ref{fig:stylizedImages}. Our method consistently outperforms all reference methods in style transfer metrics (FID, LPIPS, ArtFID) across all datasets while achieving the best or second best reconstruction scores (PSNR, SSIM). This indicates that our approach successfully applies diverse styles to training samples without compromising anatomical structure, an essential requirement for medical image augmentation. In contrast, the reference methods either fail to reliably transfer style or introduce severe anatomical distortions.
\subsection{Disease Classification}
\label{subsec:diseaseClassification}
\begin{table}[tb]
\caption{Test accuracy (\%) of a DenseNet121 disease classifier averaged over three runs per dataset. The Baseline corresponds to training with the optimal dataset-specific conventional augmentation strategy. Each subsequent row shows the result when additionally employing the respective style transfer method. The \textbf{best-performing method} per dataset is highlighted in bold.}
\label{tab:diseaseClassification}
    \centering
    \setlength{\tabcolsep}{2.4mm}
    \begin{tabular}{l r r r}
        \toprule
        \multirow{2.25}{*}{Method} & \multicolumn{1}{c}{Camelyon17} & \multicolumn{1}{c}{Epithelium} & \multicolumn{1}{c}{Fitzpatrick} \\
        & \multicolumn{1}{c}{WILDS} & \multicolumn{1}{c}{Stroma} & \multicolumn{1}{c}{17k} \\
        \midrule
        Baseline & 90.24$\pm$2.4 & 76.09$\pm$\phantom{1}3.3 & 81.18$\pm$1.1 \\
        \; + AdaIN & 93.36$\pm$1.1 & 66.23$\pm$20.8 & \textbf{81.52$\pm$0.3} \\
        \; + IEContrAST & 95.24$\pm$1.1 & 63.40$\pm$20.2 & 80.58$\pm$0.4 \\
        \; + SANet & 95.62$\pm$0.3 & 58.77$\pm$13.5 & 81.33$\pm$0.4 \\
        \; + StyTR2 & 93.86$\pm$0.4 & 69.67$\pm$\phantom{1}3.3 & 80.89$\pm$0.3 \\
        \; + SGViT & 95.25$\pm$0.5 & 58.19$\pm$\phantom{1}9.7 & 80.40$\pm$0.6 \\
        \midrule
        \; + Stylizing ViT & \textbf{95.65$\pm$0.9} & \textbf{89.07$\pm$\phantom{1}0.7} & 80.92$\pm$0.6 \\
        \bottomrule
    \end{tabular}
\end{table}
We evaluate the effectiveness of our method for domain generalization and its compatibility with traditional augmentation strategies by integrating it into the training pipeline of a DenseNet121 disease classifier. The core intuition is that \textit{Stylizing ViT} should complement rather than replace established augmentation strategies. These conventional techniques address common variations in scale, color, and illumination, while our approach enhances domain generalization by generating anatomically consistent yet stylistically diverse samples. This encourages the model to prioritize structural representations over domain-specific appearance cues, a property particularly beneficial in histopathology and dermatology, where staining differences (Camelyon17-WILDS, Epithelium-Stroma) and skin tone variations (Fitzpatrick17k) often impede cross-domain robustness.

Following the WILDS protocol~\cite{wilds2021}, we report test accuracy at a fixed threshold of 0.5. We compare \textit{Stylizing ViT} against all previously introduced style transfer methods across three datasets, each representing a distinct generalization challenge. 
For each dataset, we first determine a baseline based on the optimal conventional augmentation strategy. We consider resized cropping, grayscale conversion, color jitter, RandAugment, and domain-specific augmentations from~\cite{disalvo2024medmnistc}. We identified color jitter (Camelyon17-WILDS, Fitzpatrick17k) and resized cropping (Epithelium-Stroma) as optimal methods for the respective datasets. We then quantify the benefit of each style transfer method, including ours, by employing them in addition to the selected optimal dataset-specific augmentation.

The DenseNet121 classifier is trained for 100 epochs and three independent runs under each augmentation configuration using AdamW with cosine annealing (initial learning rate 0.001), batch size 256, and early stopping. Dataset-specific augmentations are applied online with default parameters, while style transfer is applied with probability 0.33 to balance exposure to stylized and original samples. All methods perform full image-to-image transfer except SGViT, which transfers a single style patch across all anatomical regions, emphasizing local texture adaptation and currently representing the state of the art on Camelyon17-WILDS.
\tablename~\ref{tab:diseaseClassification} summarizes the results. \textit{Stylizing ViT} achieves state-of-the-art performance on Camelyon17-WILDS and yields a substantial (+\qty{13}{\%}) improvement over the next-best method on Epithelium-Stroma. On Fitzpatrick17k, however, none of the evaluated methods produce consistent gains, indicating that this dataset presents challenges that training-time augmentation alone may not adequately address.
\subsection{Test-Time Augmentation}
\label{subsec:testTimeAugmentation}
\begin{figure}[tb]
    \centering
    \includegraphics[width=0.84\linewidth]{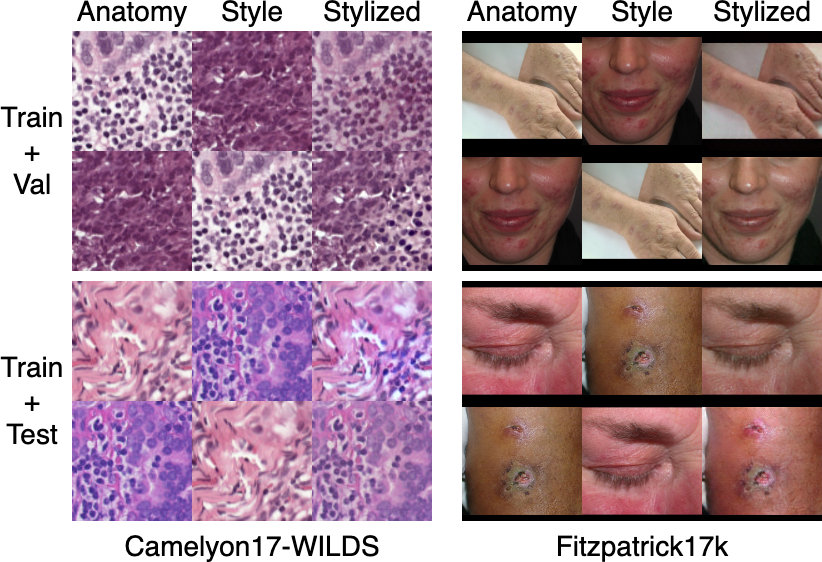}
    \caption{Style transfer results for Camelyon17-WILDS (left) and Fitzpatrick17k (right), showing high quality images across different dataset splits and domains such as staining variations and skin tones.}
    \label{fig:ablationStylizedImages}
\end{figure}
To evaluate the generalization capabilities of our method beyond the anatomies and styles encountered during training, we explore its application for augmenting validation and test images, potentially containing styles never seen by the encoder during training. Specifically, we interchange anatomy and style images across all three dataset splits. \figurename~\ref{fig:ablationStylizedImages} presents stylized results for Camelyon17-WILDS and Fitzpatrick17k, demonstrating high-quality style transfer across all splits.
These findings highlight the potential of our method for test-time augmentation, where test images are augmented with training styles to better align them with the training distribution. We quantitatively evaluate the applicability of our method for this on Camelyon17-WILDS by stylizing each test image with the style of a randomly selected training image. The stylized images are then classified using the DenseNet121 model from before trained without any data augmentation.
Our results indicate a marginally significant improvement (paired \textit{t}-test, $p < 0.1$) despite the small sample size (3 runs) and large variance in the baseline performance, increasing accuracy from \qty{59.64}{\%} to \qty{77.40}{\%}. These findings emphasize the effectiveness of our approach to enhance generalization at test time. 
\section{Discussion and Conclusion}
We introduced \textit{Stylizing ViT}, a novel Vision Transformer architecture that unifies self- and cross-attention within a single block, enabling anatomically consistent style transfer for visual domains. Our experiments show that this mechanism can be effectively utilized for data augmentation, improving robustness to domain shifts. Furthermore, \textit{Stylizing ViT} enhances domain generalization at test time by aligning unseen inputs more closely with the training distribution.

While training can be computationally demanding, primarily due to the joint attention mechanism, our approach remains practical in real-world settings. Our stylization operates within milliseconds per batch and can be seamlessly integrated into the data loading pipeline, making it well suited for efficient, on-the-fly augmentation during training.
In terms of performance, \textit{Stylizing ViT} achieves state-of-the-art results on Camelyon17-WILDS and substantial improvements on Epithelium-Stroma. Although its impact is more limited on Fitzpatrick17k, qualitative results remain convincing. This suggests that datasets involving complex, subtle, or patient-specific domain shifts, such as variations in skin tone, lighting, or photographic conditions, may require additional strategies.
\section{Compliance with ethical standards}
\label{sec:ethics}

This research study was conducted retrospectively using human subject data made available in open access by~\cite{wilds2021},~\cite{Beck2011},~\cite{Linder2012}, and~\cite{Groh2021}. Ethical approval was not required as confirmed by the license attached with the open access data.

\section{Acknowledgments}
\label{sec:acknowledgments}

This work was funded through the Hightech Agenda Bayern (HTA) of the Free State of Bavaria, Germany.

\bibliographystyle{IEEEbib}
\bibliography{strings,refs}

%
\appendix
\section{Overview of Supplementary Material}
The following sections provide additional experimental details, analyses, and qualitative results that complement the main paper. 
In Section \ref{sec:experimentalDetails}, we report implementation details, including computational setup, sources of randomness, data augmentation protocols, and reference style transfer methods used throughout our experiments. 
In Section \ref{sec:anatomyPreserving}, we illustrate the ability of the proposed method to modify visual appearance while maintaining anatomical structure via a controlled experimental setup.
In Section \ref{sec:ablationStudy}, we provide an extensive ablation study analyzing the contribution of individual loss terms, architectural components, model size, and the proportion of augmented samples used during downstream training. 
Finally, in Section \ref{sec:additionalStyleTransferResults}, we include additional qualitative style transfer results across all evaluated datasets and style transfer methods.
\section{Experimental details}
\label{sec:experimentalDetails}
\subsection{Computation}
All experiments were conducted on a single NVIDIA RTX 6000 Ada GPU (48 GB RAM) running Ubuntu 24.04.2 LTS with Python 3.12.

\subsection{Randomness}
Each result reports the average over three independent runs with different random seeds: $S \in \{265017005, 145288287, 206368124\}$

\subsection{Data Augmentation}
Traditional augmentations (random resized crop, grayscale, and color jitter) were applied using Torchvision’s built-in implementations with defaul hyperparameters. For RandAugment and the domain-specific MedMNIST-C augmentations, we used their official implementations with default hyperparameters.

\subsection{Style Transfer Reference Methods}
All reference style transfer methods were adapted from their official implementations and trained using their default settings and hyperparameters.
\section{Anatomy-Preserving Style Transfer}
\label{sec:anatomyPreserving}
\begin{figure}[!htb]
    \centering
    \includegraphics[width=0.85\linewidth]{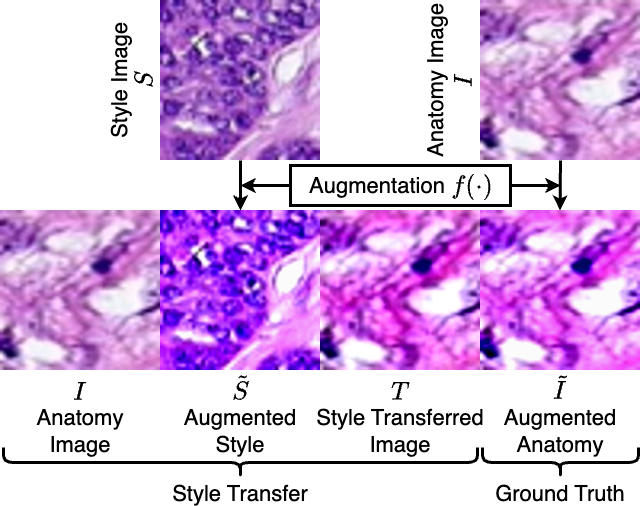}
    \caption{Demonstration of our method’s ability to transfer style while preserving anatomical structure. Given an input image pair $(I, S)$, we first apply a shared color jitter augmentation $f(\cdot)$ to produce $\Tilde{I}$ and $\Tilde{S}$. Our model generates $T$, which closely matches $\Tilde{I}$, indicating successful transfer of style while maintaining anatomical fidelity.}
    \label{fig:poc}
\end{figure}
We assess whether our method can transfer the style of a source image $\Tilde{S}$ onto a target image $I$ while preserving the anatomical content of $I$. Since style lacks a precise definition in images, we construct a controlled evaluation protocol using a predefined augmentation function $f(\cdot)$ that alters style while preserving content. Specifically, we define a color-altering transform $f(X)$, which converts an image $X$ into $\Tilde{X}$. This controlled setup allows us to create both the style reference $\Tilde{S} = f(S)$ and target image $\Tilde{I} = f(I)$ for evaluating the stylization of $I$ in the style of $\Tilde{S}$.

\figurename~\ref{fig:poc} presents this evaluation for a single image pair $(I, S)$ from the Epithelium-Stroma dataset. We apply a color jitter augmentation $f$ to generate the style reference $\Tilde{S}$ and the expected target $\Tilde{I}$. We then apply our method to transfer the style from $\Tilde{S}$ to $I$, resulting in the stylized output image $T$. $T$ visually aligns with the expected target $\Tilde{I}$, indicating that our method accurately transfers the style of $\Tilde{S}$ without distorting the anatomy of $I$.

\begin{figure}[htb]
    \centering
    \includegraphics[width=0.9\linewidth]{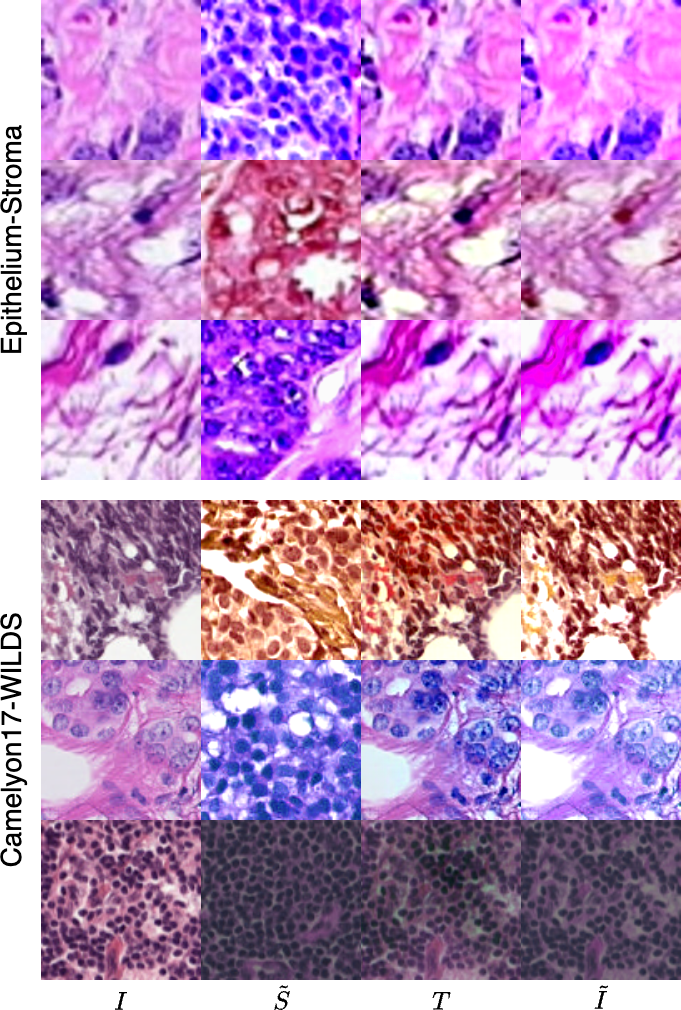}
    \caption{Additional controlled style transfer examples from the Epithelium-Stroma and Camelyon17-WILDS datasets. For each row, from left to right: (1) anatomy image $I$, (2) augmented style image $\tilde{S} = f(S)$, (3) stylized output $T$, and (4) expected target $\tilde{I} = f(I)$. The style transform $f$ corresponds to PyTorch’s \texttt{ColorJitter} augmentation with default parameters.}
    \label{fig:poc_supplementary}
\end{figure}

Additional qualitative examples from both the Epithelium-Stroma and Camelyon17-WILDS datasets are provided in \figurename~\ref{fig:poc_supplementary}.
\section{Ablation Study}
\label{sec:ablationStudy}
We conduct an ablation study on the Epithelium-Stroma dataset to quantify the contribution of individual loss terms and model components. For each configuration, we retrain the model from scratch and evaluate both reconstruction quality (using identical input images) and style transfer quality (using distinct anatomy and style images). Reconstruction quality is assessed using PSNR and SSIM, while style transfer quality is evaluated with FID and ArtFID.

In addition, we study the impact of encoder size across all three datasets and analyze how the proportion of stylized samples used during downstream classifier training affects classification performance.
\subsection{Loss Terms}
\label{subsec:lossTerms}
\begin{table}[htb]
    \caption{Impact of individual loss terms on reconstruction and style transfer quality. PSNR and SSIM ($\shortuparrow$) measure reconstruction performance while FID and ArtFID ($\shortdownarrow$) assess style transfer quality.}
    \label{tab:ablationLoss}
    \centering
    \begin{tabular}{l r r r r}
        \toprule
        \multirow{2.5}{*}{Method} & \multicolumn{2}{c}{Reconstruction} & \multicolumn{2}{c}{Style Transfer} \\
        \cmidrule(lr){2-3} \cmidrule(lr){4-5}
        & PSNR $\shortuparrow$ & SSIM $\shortuparrow$ & FID $\shortdownarrow$ & ArtFID $\shortdownarrow$ \\
        \midrule
        Identity         & 39.5 & 0.99 & 35.7 & 36.8 \\
        \; + Consistency & 40.8 & 0.99 & 35.0 & 36.1 \\
        \; + Anatomy     & 39.5 & 0.98 & 33.7 & 35.0 \\
        \; + Style       & 37.7 & 0.98 & 28.9 & 31.4 \\
        \bottomrule
    \end{tabular}
\end{table}
To analyze the effect of each loss term, we progressively introduce components into the full objective during training. Starting with only the identity loss $\mathcal{L}_{\text{i}}$, we incrementally add the consistency loss $\mathcal{L}_{\text{c}}$, anatomy loss $\mathcal{L}_{\text{a}}$, and finally the style loss $\mathcal{L}_{\text{s}}$.

\begin{figure}[htb]
    \centering
    \includegraphics[width=\linewidth]{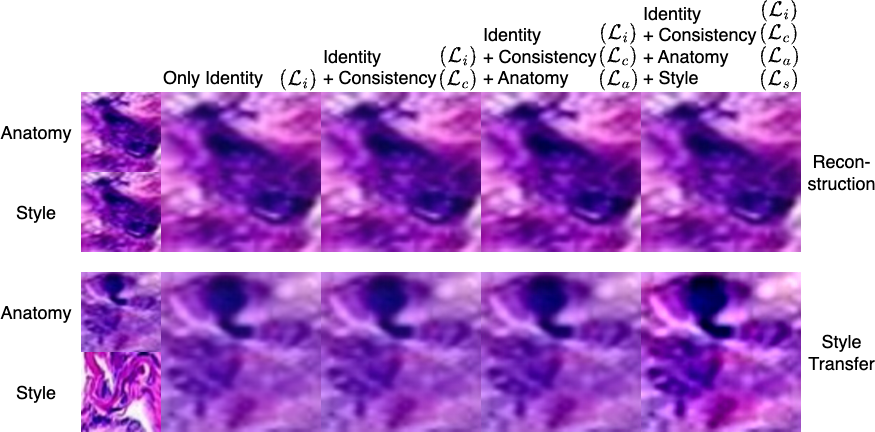}
    \caption{Qualitative evaluation of reconstruction and style transfer results on Epithelium-Stroma under different loss configurations.}
    \label{fig:ablationLoss}
\end{figure}

Reconstruction performance is evaluated on the training set, while style transfer is assessed using the validation set as anatomy input and the training set as style input. This pairing avoids over-emphasizing trivial solutions. Quantitative results are reported in \tablename~\ref{tab:ablationLoss}, with representative qualitative examples shown in \figurename~\ref{fig:ablationLoss}.

The results indicate that the identity and consistency losses primarily drive high-fidelity reconstruction. Incorporating the anatomy loss improves the preservation of fine structural details during style transfer, while the style loss is essential for enabling meaningful stylistic transformation. In its absence, the model tends to revert to reconstructing the anatomy image without effective style modulation.

\subsection{Model Components}
\label{subsec:modelComponents}
\begin{figure}[htb]
    \centering
    \includegraphics[width=\linewidth]{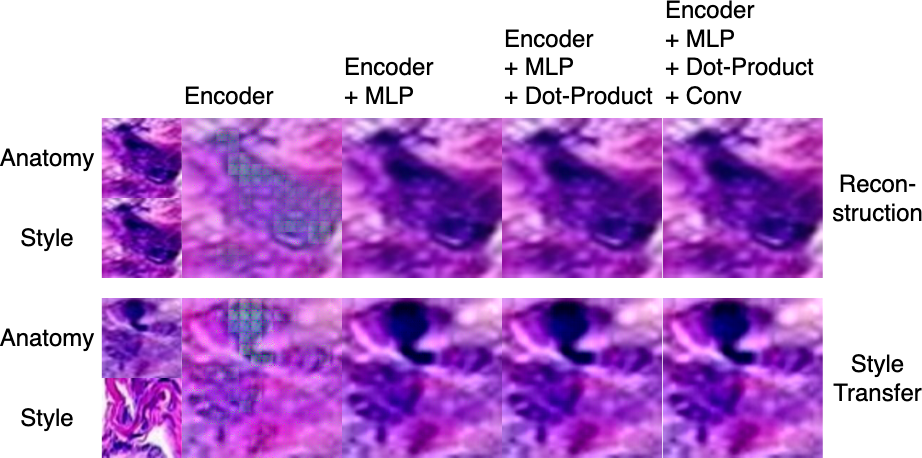}
    \caption{Qualitative evaluation of reconstruction and style transfer results on Epithelium-Stroma for different model configurations.}
    \label{fig:ablationModelParts}
\end{figure}
We further investigate the contribution of each architectural component. Starting from the encoder alone, we incrementally add: (i) the MLP; (ii) the dot-product-based reconstruction; and (iii) the final convolution. Both reconstruction and style transfer performance are evaluated on the training set. Qualitative comparisons are shown in \figurename~\ref{fig:ablationModelParts}, with quantitative results summarized in \tablename~\ref{tab:ablationModelParts}.

The results reveal that both the MLP and dot-product are crucial for accurate reconstructions. While the dot-product mechanism slightly degrades style transfer performance, it enhances the preservation of anatomical details. Finally, the inclusion of the convolutional refinement layer leads to improvements in both reconstruction and style transfer quality.
\begin{table}[htb]
    \caption{Impact of individual model components on reconstruction and style transfer quality. PSNR and SSIM ($\shortuparrow$) measure reconstruction performance while FID and ArtFID ($\shortdownarrow$) assess style transfer quality.}
    \label{tab:ablationModelParts}
    \centering
    \begin{tabular}{l r r r r }
        \toprule
        \multirow{2.5}{*}{Method} & \multicolumn{2}{c}{Reconstruction} & \multicolumn{2}{c}{Style Transfer} \\
        \cmidrule(lr){2-3} \cmidrule(lr){4-5}
        & PSNR $\shortuparrow$ & SSIM $\shortuparrow$ & FID $\shortdownarrow$ & ArtFID $\shortdownarrow$ \\
        \midrule
        Encoder & 25.7 & 0.78 & 49.1 & 65.9 \\
        \; + MLP & 35.8 & 0.98 & 2.1 & 3.2 \\
        \; + Dot-Product & 36.4 & 0.98 & 2.6 & 3.7 \\
        \; + Conv-layer & 39.0 & 0.99 & 1.5 & 2.6 \\
        \bottomrule
    \end{tabular}
\end{table}
\subsection{Model Size}
\label{subsec:modelSize}
\begin{figure}[htb]
    \centering
    \includegraphics[width=\linewidth]{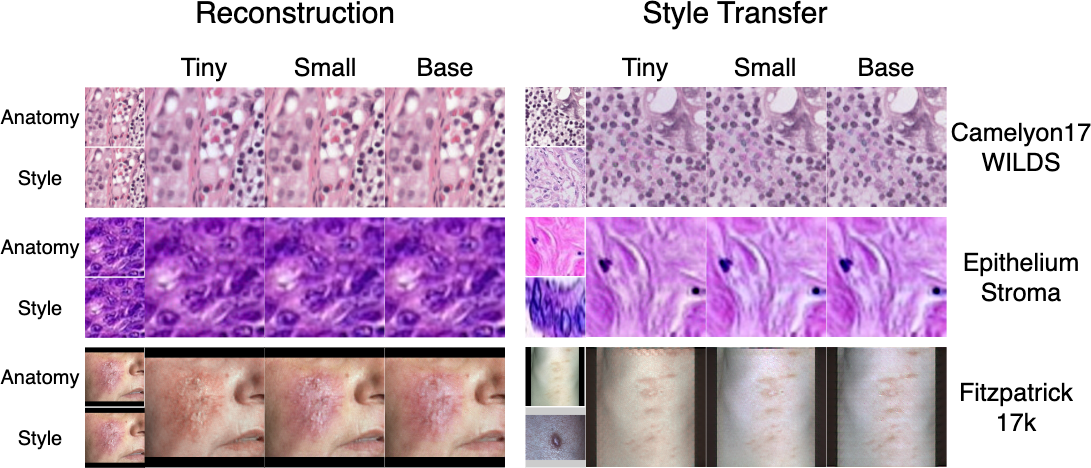}
    \caption{Qualitative comparison of reconstruction and style transfer results across encoder sizes (tiny, small, base) for exemplary images of Camelyon17-WILDS, Epithelium-Stroma, and Fitzpatrick17k.}    
    \label{fig:ablationModelSize}
\end{figure}
%


We examine the effect of encoder capacity on reconstruction and style transfer performance by replacing the \textit{Stylizing ViT} encoder in its base configuration with two smaller variants: a \emph{tiny} model (3 attention heads, 192-dimensional embeddings) and a \emph{small} model (6 attention heads, 384-dimensional embeddings).

Qualitative and quantitative results across datasets (\figurename~\ref{fig:ablationModelSize} and \tablename~\ref{tab:ablationModelSize1}) indicate that the tiny variant consistently underperforms the larger models in both reconstruction and stylization tasks. In contrast, the small variant provides a competitive alternative to the base configuration, particularly on smaller datasets such as Epithelium-Stroma and Fitzpatrick17k. Notably, on Fitzpatrick17k, the small model slightly outperforms the base model in terms of style transfer quality.

On Camelyon17-WILDS, however, the base configuration consistently yields the best overall performance, suggesting that higher model capacity is beneficial when sufficient training data are available. Overall, these results demonstrate the flexibility of the proposed framework, which can be adapted to different dataset sizes and computational constraints. In particular, the small variant offers a favorable trade-off between performance and efficiency, making it suited for deployment in resource-constrained settings.

\begin{table}[htb]
\caption{Comparison of reconstruction (PSNR, SSIM) and style transfer (FID, LPIPS, and ArtFID) quality of tiny, small, and base \textit{Stylizing ViT} on the training sets of Camelyon17-WILDS, Epithelium-Stroma, and Fitzpatrick17k.}
\label{tab:ablationModelSize1}
    \centering
    \setlength{\tabcolsep}{1.9mm}
    \begin{tabular}{l r r r r r}
        \toprule
        \multirow{2.5}{*}{Backbone} & \multicolumn{2}{c}{Reconstruction} & \multicolumn{3}{c}{Style Transfer} \\
        \cmidrule(lr){2-3} \cmidrule(lr){4-6}
        & PSNR $\shortuparrow$ & SSIM $\shortuparrow$ & FID $\shortdownarrow$ & LPIPS $\shortdownarrow$ & ArtFID $\shortdownarrow$ \\
        \midrule
        \multicolumn{5}{l}{\textit{Camelyon17-WILDS}} \\
        \:\:\: Tiny  & 33.4 & 0.95 & 10.6 & 0.08 & 12.5 \\
        \:\:\: Small & 41.5 & 0.98 & 6.4 & 0.06 & 7.9 \\
        \:\:\: Base  & 45.4 & 0.99 & 6.2 & 0.06 & 7.6 \\
        \midrule
        \multicolumn{5}{l}{\textit{Epithelium-Stroma}} \\
        \:\:\: Tiny  & 30.7 & 0.96 & 19.4 & 0.04 & 21.1 \\
        \:\:\: Small & 38.9 & 0.99 & 1.5  & 0.02 & 2.6 \\
        \:\:\: Base  & 39.0 & 0.99 & 1.5  & 0.02 & 2.6  \\
        \midrule
        \multicolumn{5}{l}{\textit{Fitzpatrick17k}} \\
        \:\:\: Tiny  & 30.7 & 0.84 & 33.2 & 0.18 & 40.5 \\
        \:\:\: Small & 30.6 & 0.82 & 21.6 & 0.16 & 26.3 \\
        \:\:\: Base  & 31.5 & 0.77 & 36.9 & 0.20 & 45.5 \\
        \bottomrule
    \end{tabular}
\end{table}
\subsection{Amount of Augmented Samples}
\label{subsec:amountOfAugmentedSamples}
\begin{figure}[htb]
    \centering
    \includegraphics[width=0.95\linewidth]{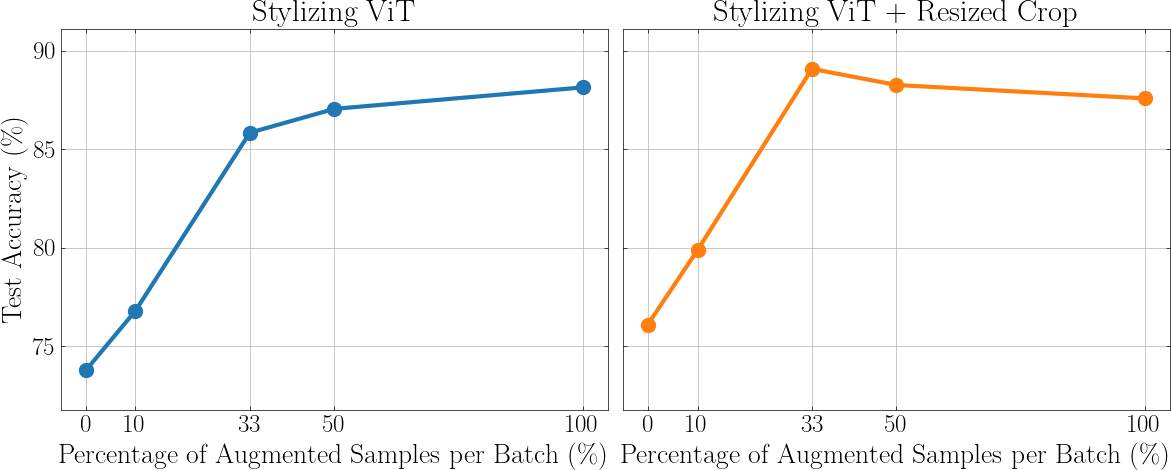}
    \caption{Impact of varying the proportion of augmented samples per batch on test accuracy for the Epithelium-Stroma dataset.
    }
    \label{fig:ablationAmountAugmentedSamples}
\end{figure}

We investigate the effect of varying the proportion of augmented samples used during downstream classifier training on the Epithelium-Stroma dataset. Specifically, we compare different augmentation ratios: 0\%, 10\%, 33\%, 50\%, and 100\%, applied per mini-batch, using stylized samples generated by our method. We evaluate two configurations: (i) pure stylization via our proposed model, and (ii) stylization combined with a resized crop augmentation applied to the style images prior to generation.
The results, visualized in \figurename~\ref{fig:ablationAmountAugmentedSamples}, indicate that the downstream test accuracy remains remarkably stable once the proportion of augmented samples reaches approximately 33\%. Notably, the 100\% augmentation setting achieves competitive and in some cases superior performance compared to more conservative configurations such as 33\% or 50\%. This suggests that our stylization framework is capable of generating sufficiently diverse and label-preserving samples, making even fully augmented batches viable for training without degrading performance.
\section{Additional Style Transfer Results}
\label{sec:additionalStyleTransferResults}
\begin{figure}[!htb]
    \centering
    \includegraphics[width=\linewidth]{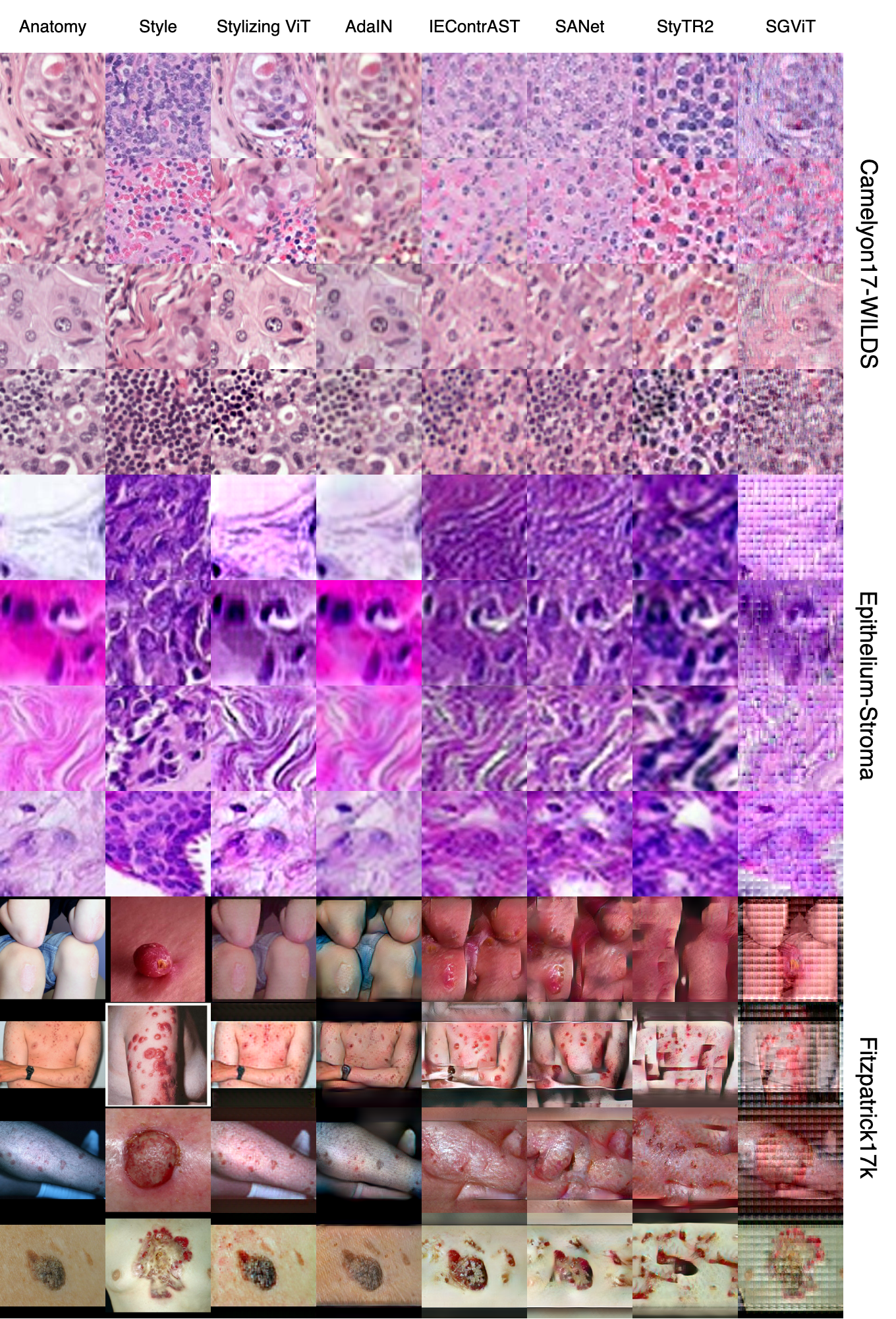}
    \caption{Additional stylized images generated by all evaluated style transfer methods on training image pairs from Camelyon17-WILDS, Epithelium-Stroma, and Fitzpatrick17k.}
    \label{fig:stylizedImagesSupplementary}
\end{figure}
%
%

\end{document}